# On Cooperative Coevolution and Global Crossover

Larry Bull and Haixia Liu

School of Computing and Creative Technologies

University of the West of England, Bristol UK

Larry.bull@uwe.ac.uk

Abstract—Cooperative coevolutionary algorithms (CCEAs) divide a given problem in to a number of subproblems and use an evolutionary algorithm to solve each subproblem. This letter is concerned with the scenario under which a single fitness measure exists. By removing the typically used subproblem partnering mechanism, it is suggested that such CCEAs can be viewed as making use of a generalised version of the global crossover operator introduced in early Evolution Strategies. Using the well-known NK model of fitness landscapes, the effects of varying aspects of global crossover with respect to the ruggedness of the underlying fitness landscape are explored. Results suggest improvements over the most widely used form of CCEAs, something further demonstrated using other well-known test functions.

Index Terms—coevolution, cooperation, crossover, multi-parent, NK model.

I. INTRODUCTION

Cooperative coevolutionary algorithms (CCEAs) divide a given task in to a number of subproblems (S) and run an evolutionary algorithm (EA) within each corresponding (sub)population. It is argued that by dividing the problem in this way, more complex problems may be amenable to solution via simulated evolution. Husbands and Mill [8] were the first to use a CCEA for problem solving, tackling a job-shop scheduling task. They divided the task such that one subproblem represented each job and there was a further arbitration subproblem, with results presented for two jobs (i.e., S=3). The problem was formulated such that each job subproblem had access to a local fitness measure and then solutions of equal rank in each subproblem were partnered to form whole solutions. The ranking step was later improved through the use of a grid-based approach [7]. Bull and Fogarty [3] used a CCEA to coevolve cooperating rule-based controllers for a very simple mobile robot task, with S=2 and later S=3 [4]. Their partnering scheme combined the simultaneously created offspring from each subproblem for a fitness evaluation. Potter and DeJong [13] applied a CCEA to function optimization with the N problem variables each evolved within a separate population, i.e., S = N. Moreover, they introduced a round-robin scheme such that each subproblem evolves in turn, where a newly created (sub)solution is partnered with the current fittest individual in the other S-1 subproblem populations (termed CCGA-1). The partnering scheme was found to be suboptimal for some functions and further extended to using the best and a random individual from each subproblem, assigning the better fitness (CCGA-2). The work is typically described as the first use of a CCEA and its round-robin scheme has been almost ubiquitously adopted.

Hence under cooperative coevolution, a traditional fitness function $f: G \rightarrow R$, where $G$ is a set of $N$ variables $\{g_1, g_2, \ldots, g_N\}$, becomes $f: C \rightarrow R$, where $C$ is a set of grouped variables $\{C_1 = \{g_1, \ldots, g_{(N/S)}\}, \ldots, C_{N/S} = \{g_{(N/S) \cdot (S-1)}, \ldots, g_N\}\}$. As first noted in [4], by dividing a task, cooperative coevolution shares some similarity with crossover: new solutions are created by combining individuals, where fixed crossover points may be seen to exist between each subproblem. Crossover in most EAs occurs between two individuals selected from a given population. That is, selection is run twice and the variable values are swapped between the two parent solutions from one or more positions. However, within Evolution Strategies (ES), Schwefel [15] introduced what he termed a global crossover operator such that selection is run once for each variable to create a new solution (e.g., see [1] for a description): selection is run $N$ times to create a new solution of $N$ variables for evaluation. This is equivalent to a CCEA using a single fitness measure and the simultaneously new offspring partnering scheme [3] with $S=N$, i.e., without round-robin evaluations wherein only one subproblem generates a new offspring per evaluation [13]. This realisation can be generalised for $0 < S \leq N$ in what are here termed global crossover EAs (GCEAs). Following [2], the effects of different partnering schemes under the round-robin approach have been explored many times (e.g., [17][16][11]). The basic properties of GCEAs, that is, CCEAs without the round-robin, are explored here using the NK model of fitness landscapes [10], as well as others.

II. THE NK MODEL

The NK model allows the systematic study of various aspects of fitness landscapes. In the standard model, the features of the fitness landscapes are specified by two parameters: $N$, the length of the genome; and $K$, the number of randomly chosen genes that has an effect on the fitness contribution of each binary gene. Thus increasing $K$ with respect to $N$ increases the epistatic linkage, i.e., the number of genes that interfere with each other in their contributions to fitness (when $K=0$ the function is separable), increasing the ruggedness of the fitness landscape. The increase in epistasis increases the number of optima, increases the steepness of their sides, and decreases their correlation [9]. The model assumes all intragenome interactions are so complex that it is only appropriate to assign random values to their effects on fitness. Therefore for each of the possible K interactions, a table of $2^{(K+1)}$ fitnesses is created, with all entries in the range 0.0 to 1.0, such that there is one fitness value for each combination of traits. The fitness contribution of each trait is found from its individual table. These fitnesses are then summed and normalised by $N$ to give the selective fitness of the individual (Figure 1).

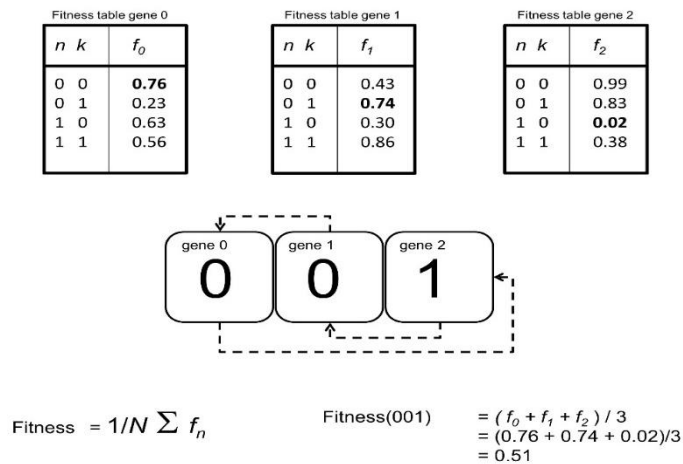

Figure 1. An example NK model where $N$=3 and $K$=1.

III. GLOBAL CROSSOVER

As noted above, Schwefel [15] introduced the concept of global crossover wherein the selection function is called each time to determine the parent donating the next variable in the creation of an offspring (see also ant colony optimization, e.g., [5]). Generalising the mechanism, $S$ refers to the number of times the selection function is called, i.e., equivalent to the number of subproblem (sub)populations in CCEAs. When $S < N$, the location of the crossover points can be chosen at random or each can be equally spaced ($N/S$). The latter approach - here termed GCEA-0 - is akin to the scheme typically used within CCEAs: starting from the left-hand most variable of the overall task representation, $S$-1 fixed global crossover positions are chosen. For the former approach – here simply termed GCEA – the global crossover process begins at the location of the first position chosen and the variables are therefore assumed to be in a ring. Moreover, the crossover positions of each global crossover process are re-chosen (at random) per offspring generation (Algorithm 1).

```
Algorithm 1 Basic global crossover evolutionary algorithm.
    initialise(pop)
    evaluate(pop)
    Repeat until terminationcriteria
        for( s = 1 to S )
            global crossover points[s] = random(N)
        sort ascending (global crossover points)
        count = global crossover points[1]
        parent = selection(pop)
        for( g = 1 to N )
            for( s = 1 to S )
                if( global crossover points[s] == g )
                    parent = selection (pop)
            offspring.genes[count] = parent.genes[count]
            count = count + 1
            if( count > N )
                count = 1
        mutate (offspring)
        evaluate (offspring)
        replace (offspring, pop)
```

## IV. EXPERIMENTATION

A simple underlying steady-state evolutionary process is used here: reproduction selection is via a binary tournament (size 2), replacement selection uses a randomly chosen individual (with elitism), the population P contains 50 individuals, genomes are binary strings of length $N=1000$ or $N=5000$, and mutation is deterministic at rate $1/N$ (one gene chosen at random for mutation). The standard EA, and as used in the CCEA (sub)populations (Algorithm 2), selects two parents and uses one-point crossover to produce a single offspring. The results presented are the average fitness of the best solution after 100,000 evaluations, for various $K$ and $S$, from 50 runs.

**Algorithm 2** Basic cooperative coevolutionary algorithm.

```
for( s = 1 to S )
    initialise(pop[s])
for( s = 1 to S )
    evaluate(pop[s], Rand)
Repeat until terminationcriteria
    for( s = 1 to S )
        parent1 = selection(pop[s])
        parent2 = selection (pop[s])
        offspring = crossover(parent1, parent2)
        mutate (offspring)
        evaluate (offspring, Best)
        replace (offspring, pop[s])
```

Figure 2 shows results for $S=2$, i.e., selection is run twice in each algorithm, with $N=1000$ and various $K$. It can be seen (a) that the GCEA gives better solutions in comparison to an equivalent EA and the GCEA-0 (T-test, $p<0.05$) for $K>4$. The GCEA-0 is not better than the EA (T-test, $p \geq 0.05$) for any $K$. The performance of the round-robin approach under the steady state scheme is also shown (b). The GCEA again gives the better solutions (T-test, $p<0.05$), here for $K>2$. The basic CCEA used here with the best partner scheme (CCEA-1, as CCGA-1 [13]) is no better than the EA regardless of K (T-test, $p \geq 0.05$) and perhaps somewhat unexpectedly with the fittest of two partners strategy (CCEA-2) performs particularly badly. Further examination (not shown) suggests this to be the case once $N>500$.

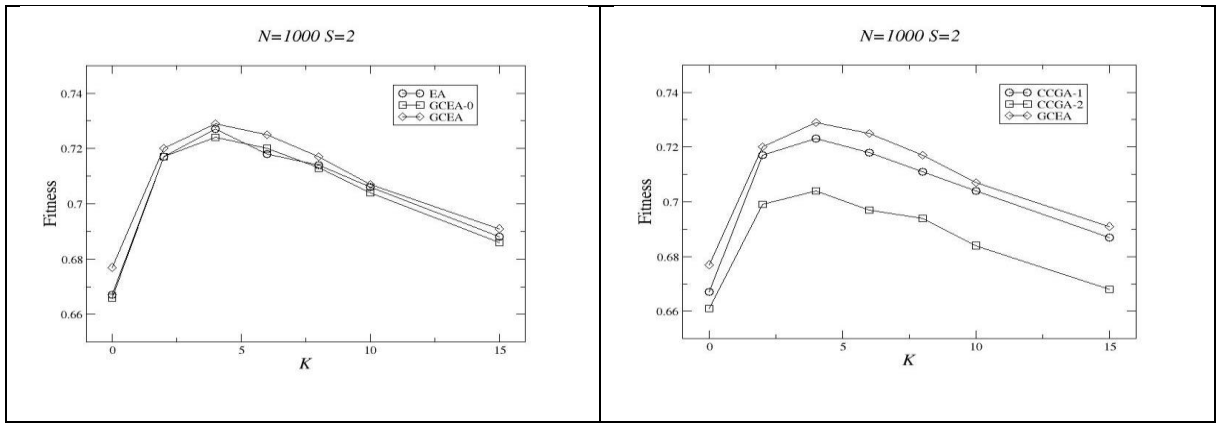

Figure 2. Showing example comparative performance of global crossover and cooperative coevolution algorithms.

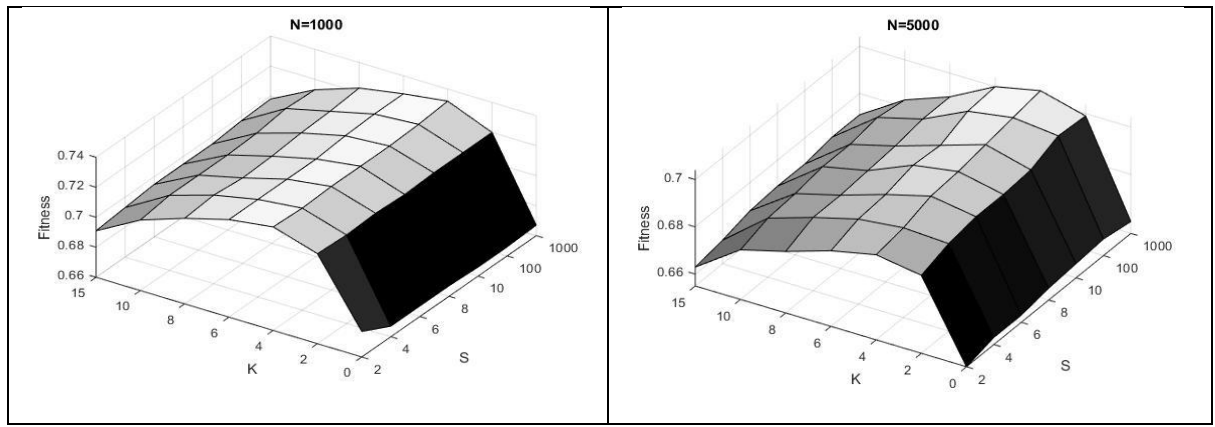

Figure 3. Showing the effects on fitness for GCEA from varying S over landscapes of different ruggedness $K$ and size $N$.

Figure 3 shows the effects of increasing $S$ for various $K$. When $N$ =1000 and $K$ > 2, fitness increases with increasing $S$, such that when $S \geq 100$ fitness is highest and higher than $S \leq 10$ (T-test, $p$ <0.05). There is no significant difference between $S$=100 and $S = N$ (T-test, $p \geq 0.05$), which is further true for $S \geq 10$ when $K \geq 8$. The same general result is also seen with $N$ =5000 but with fitnesses increasing with $S$ when $K$ >0.

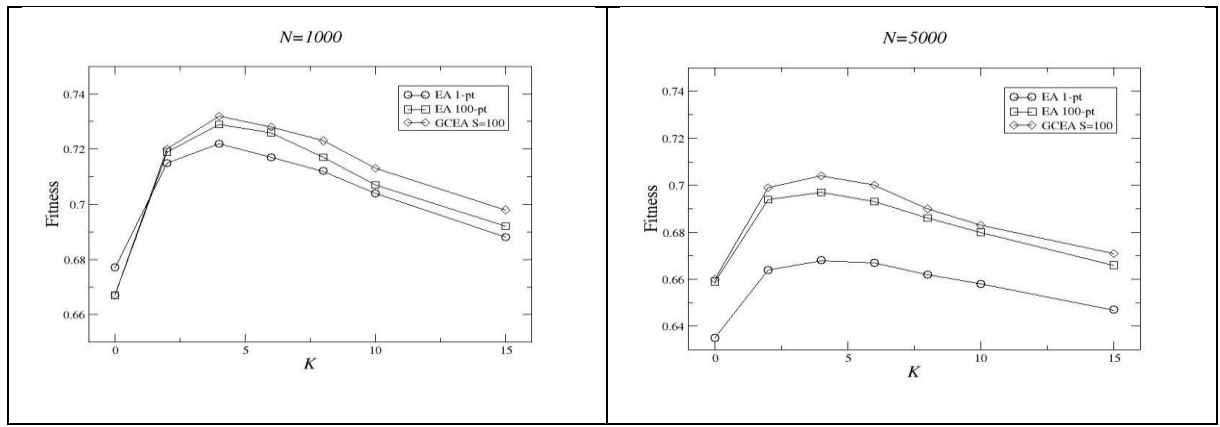

Figure 4. Showing example comparative performance of the best global crossover EA and a standard EA with the equivalent number of crossover points.

Figure 4 compares a standard EA with the same number of crossover points to those of the best global crossover EA: fitness is higher with the GCEA for all $K > 6$ (T-test, $p < 0.05$) than with two-parents using 100-point crossover with $N = 1000$ (a) and for all $K > 2$ (T-test, $p < 0.05$) with $N = 5000$ (b).

Finally, Figure 5 shows the comparative performance on some other well-known separable (Rastrigin, Sphere) and partially overlapping (Rosenbrock, Dixon-Price) minimisation functions. That is, as a contrast to the binary NK functions of variable separability used above. The effects of varying $N$ with $S=2$ is shown. All experimental details remain the same as for the NK model but under mutation a gene is altered uniformly random from the range [-1.0, 1.0]. As can be seen, the GCEA gives the best solutions when $N > 100$ (T-test, $p < 0.05$) for these functions. Increasing $S$ generally improves fitness, e.g., $S=N$ is better than $S=100$ (T-test, $p < 0.05$) for these functions, as it did for the larger NK models (not shown).

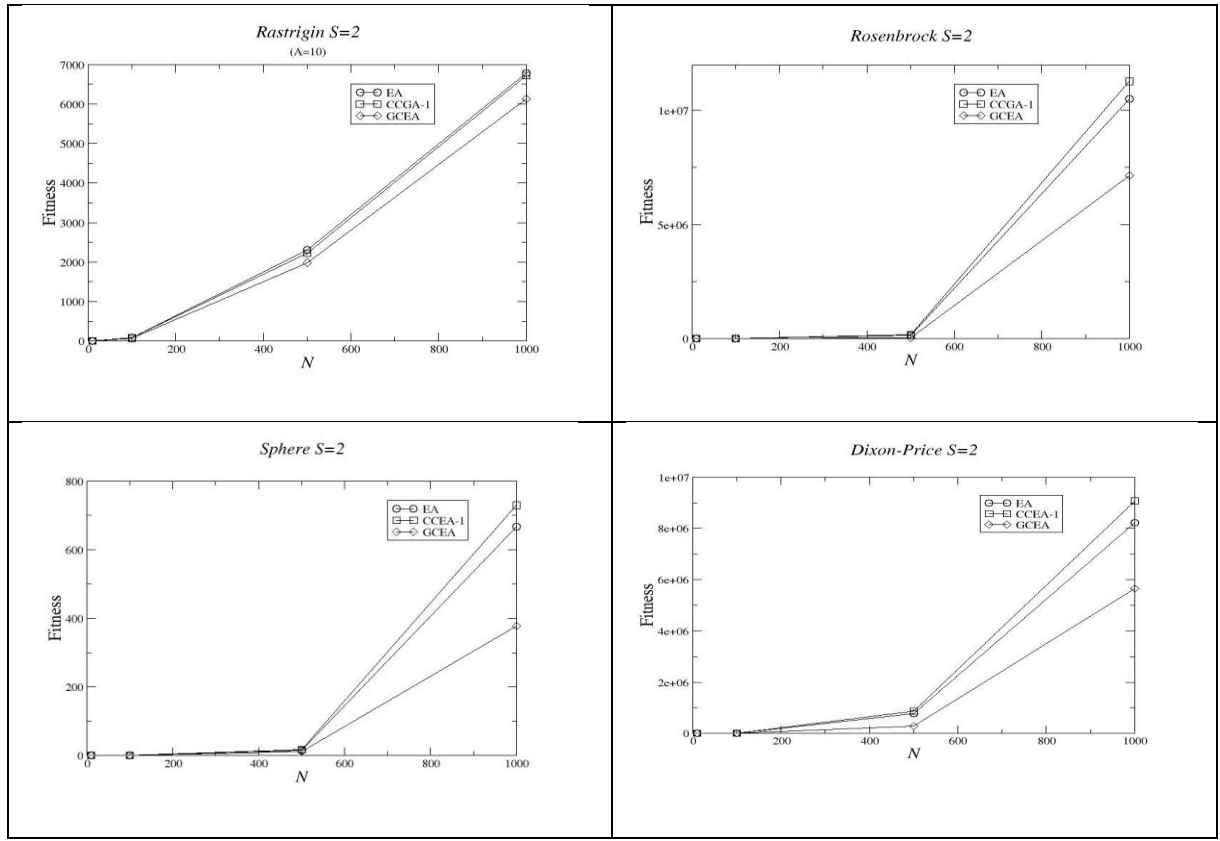

Figure 5. Showing example comparative performance of global crossover and cooperative coevolution algorithms on well-known benchmark functions.

## V. CONCLUSION

This letter has drawn attention to the similarities between global crossover as introduced in ESs and cooperative co-evolutionary EAs when a single fitness measure exists and the latter are not using a round-robin scheme. Results show no improved performance from the GCEA compared to the CCEA on smaller and/or less rugged fitness landscapes but benefits are seen thereafter. Whether the CCEA would improve in relative performance with an increase of $S$ times the number of evaluations thereby enabling it to create the same number of whole new solutions as the GCEA has not been considered.

From a cooperative coevolution perspective, although not explored here, introducing a probability $p_S$ rather than a fixed $S$ would enable the dynamic variation in the number of subproblems around a mean. Moreover, global crossover is a simple way by which to dynamically vary the size of subproblems thereby enabling a form of random variable grouping within CCEAs. Future work will compare GCEAs to CCEAs with more complex grouping methods included since the round-robin scheme has typically been used to date (e.g., see [12] for a comprehensive review). Other known uses of multiple parents ($p$) typically consider $1 < p \leq P$ parents and introduce novel crossover operators (after [14]). That is, the number of parents and crossover points is decoupled. The approach used here can be seen as a version of Eiben et al.'s [6] fitness-based scanning crossover. Future work will also compare with other multiple parent methods.